\documentclass[runningheads]{llncs}
\usepackage[T1]{fontenc}
\usepackage{graphicx,verbatim}
\usepackage{amsmath}
\usepackage{amsfonts}
\usepackage{multirow}
\usepackage[table]{xcolor}
\usepackage{booktabs}
\usepackage{amssymb} % for \checkmark
\usepackage{booktabs}
\usepackage{graphicx}
\usepackage{xcolor}
\usepackage{marvosym}
\begin{document}
\title{Frequency-enhanced Multi-granularity Context Network for Efficient Vertebrae Segmentation}
%
% \begin{comment}
\author{Jian Shi\inst{1},
Tianqi You \inst{1},
Pingping Zhang\inst{2}\textsuperscript{(\Letter)},
Hongli Zhang\inst{3}, Rui Xu\inst{1}, and Haojie Li\inst{4}}

\authorrunning{Jian Shi et al.}
\institute{School of Software Technology \& DUT-RU International School of Information Science and Engineering, Dalian University of Technology, Dalian, China
\and
School of Future Technology, School of Artificial Intelligence, Dalian University of Technology, Dalian, China
\and
	The Fifth Affiliated Hospital of Zhengzhou University, Zhengzhou, China
\and
College of Computer Science and Engineering, Shandong University of Science and Technology, Qingdao, China
\\
\email{zhpp@dlut.edu.cn}
}

% \end{comment}

% \author{Anonymized Authors}  %% Added for anonymized MICCAI 2025 submission
% \authorrunning{Anonymized Author et al.}
% \institute{Anonymized Affiliations \\
%     \email{email@anonymized.com}}

\maketitle              % typeset the header of the contribution
\begin{abstract}
Automated and accurate segmentation of individual vertebra in 3D CT and MRI images is essential for various clinical applications.
Due to the limitations of current imaging techniques and the complexity of spinal structures, existing methods still struggle with reducing the impact of image blurring and distinguishing similar vertebrae.
To alleviate these issues, we introduce a Frequency-enhanced Multi-granularity Context Network (FMC-Net) to improve the accuracy of vertebrae segmentation.
Specifically, we first apply wavelet transform for lossless downsampling to reduce the feature distortion in blurred images.
The decomposed high and low-frequency components are then processed separately.
For the high-frequency components, we apply a High-frequency Feature Refinement (HFR) to amplify the prominence of key features and filter out noises, restoring fine-grained details in blurred images.
For the low-frequency components, we use a Multi-granularity State Space Model (MG-SSM) to aggregate feature representations with different receptive fields, extracting spatially-varying contexts while capturing long-range dependencies with linear complexity.
The utilization of multi-granularity contexts is essential for distinguishing similar vertebrae and improving segmentation accuracy.
Extensive experiments demonstrate that our method outperforms state-of-the-art approaches on both CT and MRI vertebrae segmentation datasets.
The source code is publicly available at https://github.com/anaanaa/FMCNet.


\end{abstract}
%
%
%-------------------------------------------------------------------------------------
\section{Introduction}
\begin{figure}[t]
\includegraphics[width=\textwidth]{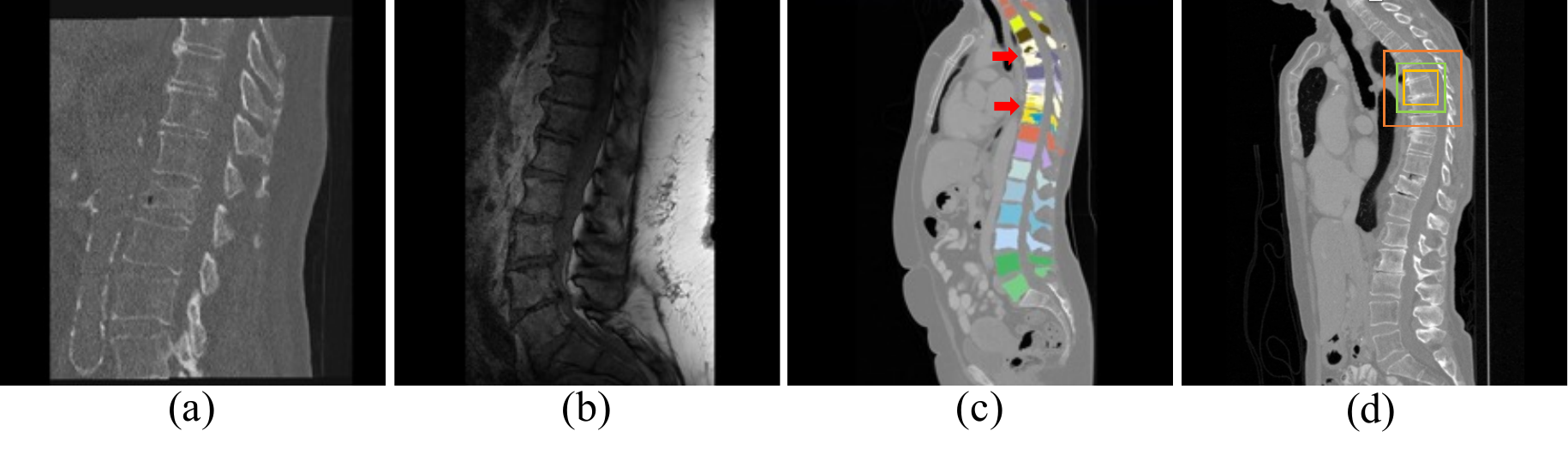}
\caption{Challenging examples in vertebrae segmentation.  (a) and (b) present examples of image blurring caused by current imaging techniques in CT and MRI modalities, respectively. (c) presents examples of the challenges in distinguishing similar vertebrae. (d) highlights the importance of multi-granularity contextual information contained in the features of different receptive fields  for distinguishing similar vertebrae.} \label{fig1}
% \vspace{-5mm}
\end{figure}

The spine is vital to the body, supporting the organ function and serving as the core of the neural pathways.
Any injury to the spine can impair neural conduction and lead to associated disorders~\cite{Wang1}. %
Automatic segmentation and identification of individual vertebra in the spine are crucial for the treatment of orthopedic diseases, as they play a key role in preoperative diagnosis, treatment planning, surgical guidance, and postoperative assessment~\cite{Pang,Lessmann,Wu}.
Although significant efforts have been made to achieve highly accurate vertebrae segmentation, the performance still remains suboptimal due to the limitations of current imaging techniques and the complexity of spinal structures~\cite{Tao}.
As illustrated in Fig.~\ref{fig1}, there are two main challenges in this task:
(1) The intrinsic properties of different 3D medical imaging modalities, such as high noise levels and low contrast in CT scans, and non-isotropic spatial resolution in MRI images, lead to image blurring and distortion.
Examples of these phenomena in CT and MRI images are in illustrated Fig.~\ref{fig1} (a) and Fig.~\ref{fig1} (b), respectively.
(2) The high similarity in shape and structure between adjacent vertebrae presents a significant challenge for accurate segmentation.
Fig.~\ref{fig1} (c) demonstrates the impact of homogeneous structures on vertebrae segmentation, showing that partial voxel predictions for adjacent vertebrae are incorrectly assigned the same label.

Existing vertebrae segmentation methods typically employ an encoder-decoder structure and have achieved commendable results~\cite{Shi1,Masuzawa}.
However, previous methods often overlook the critical impact of downsampling techniques.
The max-pooling and bilinear interpolation are commonly utilized, frequently resulting in a substantial loss of fine-grained details during the downsampling process.
This loss is especially harmful for blurred images, as retaining fine-grained details is essential for accurate segmentation.
Besides, vertebrae segmentation requires sufficient contextual information to distinguish vertebrae.
Existing Convolutional Neural Networks (CNNs) rely on local convolutional kernels to extract features, which limits their ability to extract global and geometric features.
Although Transformer-based methods~\cite{You1,You2} excel in capturing long-range dependencies, their self-attention mechanism has a high computational complexity.
Recently, the State Space Model (SSM)~\cite{Smith,Gu1} could capture the long-range dependencies with linear complexity, providing a promising solution to the limitations of CNNs and Transformers. %
While SSM lacks consideration of spatially-varying contexts, this limitation reduces its sensitivity to multi-scale contextual information in similar vertebrae, impairing its ability to distinguish them.

To mitigate the aforementioned issues, we introduce a Frequency-enhanced Multi-granularity Context Network (FMC-Net) for accurate vertebrae segmentation.
To begin with, we replace traditional downsampling methods with wavelet transform to reduce the loss of detailed information in blurred images.
The wavelet transform achieves downsampling by decomposing the features into high and low-frequency components, which are then processed separately in the subsequent steps.
High-frequency components contain a wealth of image details, textures, and edge information, but they also include noises.
Therefore, we employ a High-frequency Feature Refinement (HFR) to amplify the prominence of key features while filtering out noises, effectively restoring details and textures in blurred images.
Low-frequency components contain the most image content, we propose a Multi-granularity State Space Model (MG-SSM) to comprehensively extract low-frequency information. Specifically, MG-SSM contains granularity-aware SSMs for spatial perception.
These SSMs extract information at different granularities to distinguish similar vertebrae.
We evaluate the proposed method on both CT and MRI vertebrae segmentation datasets.
Extensive experimental results demonstrate that our approach achieves state-of-the-art performance.

Our main contributions can be summarized as follows:

$\bullet$
We propose FMC-Net that leverages wavelet transform to avoid information loss and enhances high and low-frequency features to address the challenges of image blurring and the high similarity of vertebral structures.

$\bullet$
For the high-frequency components, we introduce the HFR to restore fine-grained details.
For the low-frequency components, we employ the MG-SSM to capture spatially-varying long-range dependencies.

$\bullet$
Experiments on VERSE2019 and LUMBAR datasets demonstrate the superior performance of FMC-Net compared to other state-of-the-art approaches.
%------------------------------------------------------------------------------------------
\section{Method}
\subsubsection{The Overall Architecture:}
\begin{figure}[t]
\includegraphics[width=\textwidth]{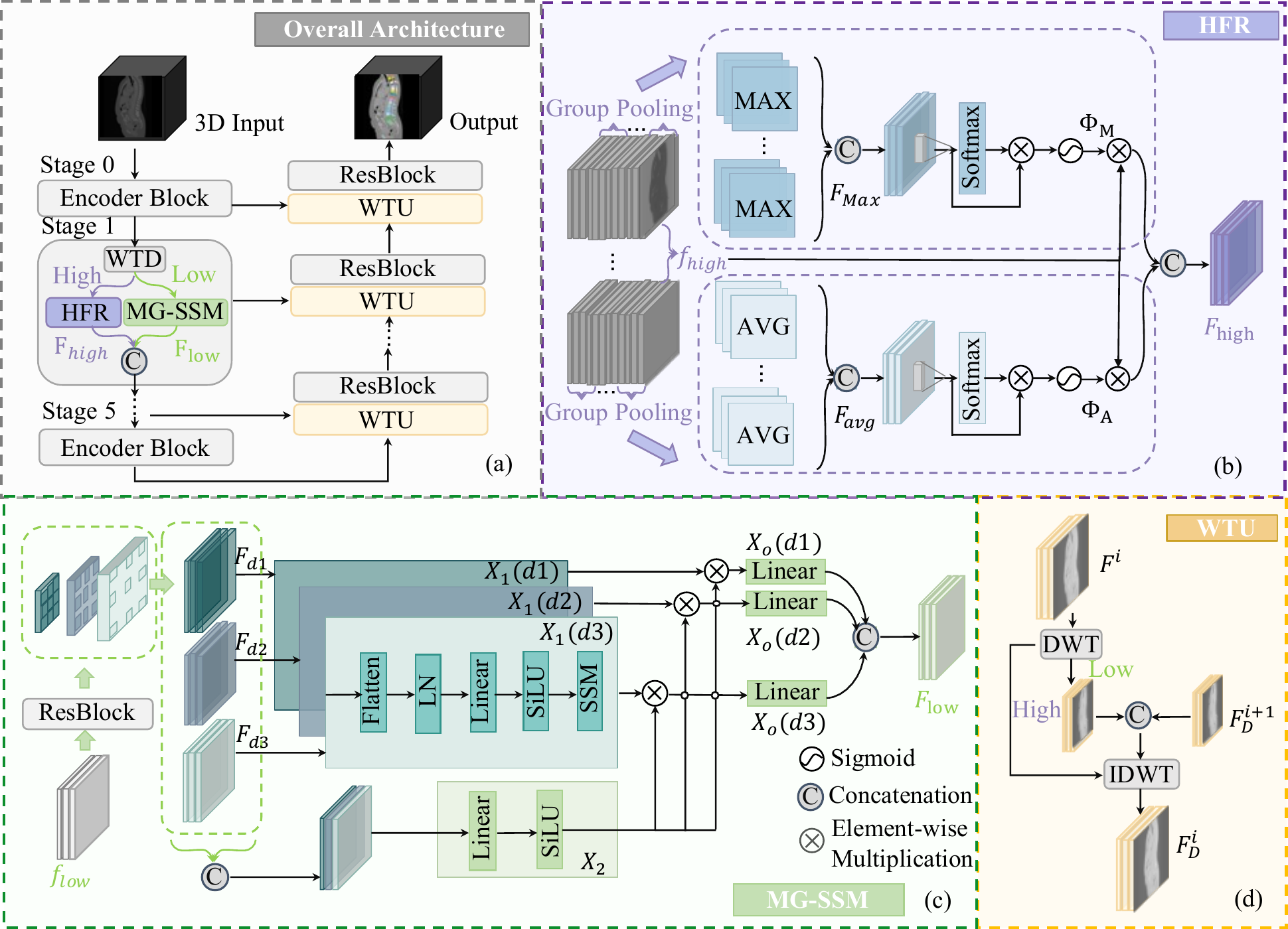}
\caption{(a) is the overall architecture of the proposed FMC-Net. (b) and (c) are HFR and MG-SSM, which enhance the high-frequency and low-frequency components, respectively. (d) is WTU, which utilizes features from different scales of the encoder while performing upsampling in the decoder.} \label{fig2}
\end{figure}

As shown in Fig.~\ref{fig2}, given a 3D input volume $I \in \mathbb{R}^{{D} \times {H} \times {W}}$, $D$, $H$ and $W$ represent the depth, height, and width of the volume, respectively.
FMC-Net aims to achieve accurate vertebrae segmentation by employing a wavelet-based encoder-decoder structure integrated with HFR and MG-SSM.
Firstly, in the encoder stage, wavelet transform enables lossless downsampling, separating the features into high-frequency and low-frequency components.
The HFR and MG-SSM are respectively applied to these components, yielding enhanced feature representations.
These enhanced components are fused as the final features at each stage, effectively capturing both semantic and detailed information.
In the decoder stage, Wavelet Transform Upsampling (WTU) employs wavelet transform-based feature fusion during upsampling, enabling the decoder to effectively leverage the information from the encoder.
%-------------------------------------------------------
%-------------------------------------------------------
\subsection{Wavelet Transform-based Sampling}
\subsubsection{Wavelet Transform Downsampling.}
Inspired by Guo \emph{et al.}~\cite{Guoping}, we implement Wavelet Transform Downsampling (WTD)  using the Discrete Wavelet Transform (DWT)~\cite{Li1} to alleviate the irreversible distortion and information loss caused by downsampling.
For the feature $F^{i} \in \mathbb{R}^{{\left(\frac{D}{2^{i}} \times \frac{H}{2^{i}} \times \frac{W}{2^{i}}\right)}}$ of the $i$-th ($ i \in (0, 1, \dots, 5)$ ) stage encoder, we employ DWT to decompose the feature $F^i$ into eight sub-wavelet bands, i.e., low-frequency component $F_{lll}^i$, and high-frequency components $F_{\{llh, \ldots, hhh\}}^i$ in horizontal, vertical, and depth dimensions (which are x-axis, y-axis and z-axis):
\begin{equation}
\left\{ F_{lll}^i, F_{llh}^i, F_{lhl}^i, F_{lhh}^i, F_{hll}^i, F_{hlh}^i, F_{hhl}^i, F_{hhh}^i, \right\} = {DWT(F^{i})},
\end{equation}
where $F^i_{\{lll, \ldots, hhh\}} \in \mathbb{R}^{\left(\frac{D}{2^{i+1}} \times \frac{H}{2^{i+1}} \times \frac{W}{2^{i+1}}\right)}$.
As low-frequency components contain the majority of the essential information, while high-frequency components capture intricate details, textures, and edge information, we focus on enhancing both of these components:
\begin{equation}
F_{low}^i = \mathcal{S}\left(F^i_{lll}\right),
\end{equation}
\begin{equation}
F_{high}^i = \mathcal{H}\left(F^i_{\{llh, \ldots, hhh\}}\right),
\end{equation}
where $F^i_{low}$ and $F^i_{high}$ denote the enhanced low-frequency and high-frequency features, respectively.
$\mathcal{S}$ denotes MG-SSM, and $\mathcal{H}$ denotes HFR.
Then the enhanced feature $F^{i+1} \in \mathbb{R}^{\left(\frac{D}{2^{i+1}} \times \frac{H}{2^{i+1}} \times \frac{W}{2^{i+1}}\right)}$ is derived by concatenating $F^i_{low}$ and $F^i_{high}$, which contains both semantic and fine-grained detail information.

\subsubsection{Wavelet Transform Upsampling.}
To fully leverage the features from the encoder, we propose Wavelet Transform Upsampling (WTU), which effectively utilizes features from different scales of the encoder while performing upsampling in the decoder.
For the features in the encoder $F^{i} \in \mathbb{R}^{{\left(\frac{D}{2^{i}} \times \frac{H}{2^{i}} \times \frac{W}{2^{i}}\right)}}$, we apply DWT to  halve the resolution and decompose them into low-frequency and high-frequency components, $F_{lll}^i$, $F_{\{llh, \ldots, hhh\}}^i \in \mathbb{R}^{\left(\frac{D}{2^{i+1}} \times \frac{H}{2^{i+1}} \times \frac{W}{2^{i+1}}\right)}$.
The low-frequency component $F_{lll}^i$ is then concatenated with the features $F_{D}^{i+1} \in \mathbb{R}^{\left(\frac{D}{2^{i+1}} \times \frac{H}{2^{i+1}} \times \frac{W}{2^{i+1}}\right)} $ from the decoder to obtain the feature $F_{fuse}$.
Finally, both the high-frequency components $F_{\{llh, \ldots, hhh\}}^i$ and $F_{fuse}$ are input into the inverse Discrete Wavelet Transform (IDWT) to generate the upsampled feature $F_{D}^{i} \in \mathbb{R}^{{\left(\frac{D}{2^{i}} \times \frac{H}{2^{i}} \times \frac{W}{2^{i}}\right)}}$.

\subsection{High-frequency Feature Refinement}
High-frequency components capture detailed textures, but also contain noises.
We employ HFR to effectively leverage high-frequency information.
The high-frequency components $f_{\text{high}}$ (comprising $F_{llh}, \ldots, F_{hhh}$) are processed through two distinct paths: one focuses on amplifying the prominence of key features, and the other filters out noises, thereby facilitating the restoration of fine-grained details and textures in blurred images.
Both paths employ spatial attention mechanisms to dynamically modulate different frequency components during training.
In the amplification path, the spatial attention mechanism $\Phi_M$ can be formulated as:
\begin{gather}
F_{\text{Max}} = \mathcal{F}_{3\times3}^{7C\rightarrow C}([\text{GroupMax}(F_{llh}), \ldots, \text{GroupMax}(F_{hhh})]), \\
\Phi_M = \sigma(\text{Softmax}(F_{\text{Max}}) \otimes F_{\text{Max}}),
\end{gather}
where $\sigma$ denotes the Sigmoid function, $[\cdot ]$ is the concatenation operation, and $\text{GroupMax}$ is the group max-pooling operation.
Each high-frequency component is divided into $2^i$ groups for pooling, and $i$ denotes the number of encoder stages.
After pooling, each component undergoes a dimension transformation that reduces the channel dimension from $2^i$ to $C$.
$\mathcal{F}^{7C\rightarrow C}_{3\times3}$ is a $3\times3$ convolutional layer that transforms the dimension of the concatenated high-frequency feature from $7C$ to $C$.
The concatenation operation aggregates the amplified features of each high-frequency component after max-pooling operations, while the Softmax operation computes the contribution of each component to the spatial attention map.
Meanwhile, the denoising path follows the same structure as the amplification path, except that the average pooling is used instead of max-pooling.
Similarly, a spatial attention map $\Phi_A$ is generated, which smooths the features of each component and effectively filters out noises.
The final enhanced high-frequency features can be obtained by the following formula:
\begin{equation}
F_{high} = [\Phi_M\otimes F_{\{llh, \ldots, hhh\}}, \Phi_A \otimes F_{\{llh, \ldots, hhh\}}].
\end{equation}

Here, the spatial attention map $\Phi_M$ is multiplied with each high-frequency component to emphasize the key features, while $\Phi_A$ performs a similar operation to filter out noises from each component.
Finally, the amplified and denoised components are concatenated to obtain the enhanced high-frequency component.

\subsection{Multi-Granularity State Space Model}
For low-frequency component $f_{low}$, we propose MG-SSM to effectively utilize the rich information for distinguishing similar vertebrae.
Firstly, we employ a residual block to extract local features.
Then we use three parallel 3D depth-wise convolutional layers with different dilation rates to capture deep features with different scales of receptive fields.
Subsequently, these features ($F_{d1}, F_{d2}, F_{d3}$) are respectively fed into three Visual State Space Modules (VSSMs) to learn multi-scale contexts.
This process can be defined as following:
\begin{equation}
\begin{aligned}
F_{dj} &= \text{DWConv}_{dj}(\text{ResBlock}(F_{lll})), j \in (1 ,2, 3), \\
X_1(dj) &= \text{SSM}\left(\phi\left(\text{Linear}\left(\text{LN} (F_{dj})\right)\right)\right),
\end{aligned}
\end{equation}
where $dj$ represents the $j$-th dilation rate, $\phi$ denotes the Sigmoid Linear Unit (SiLU) activation function~\cite{Elfwing}, and $X_1(dj)$ is the first branch of each VSSM.
Next, $F_{d1}$, $F_{d2}$, and $F_{d3}$ are concatenated to generate multi-scale features while simultaneously capturing the inter-scale relationships.
And the concatenated feature is then fed into the second branch of VSSM:
\begin{equation}
X_2 = \phi(\text{Conv}([F_{d1}, F_{d2}, F_{d3}]),
\end{equation}

The features from both branches are combined to produce the output:
\begin{equation}
X_{o}(dj) = \text{Linear}(X_1 (dj)\otimes X_2),
\end{equation}
where $X_2$ is shared among three VSSMs.
Finally, each VSSM output is passed through a linear layer, followed by concatenation to generate the final enhanced low-frequency component.
MG-SSM addresses the limitations of a single SSM in capturing spatially-varying features and provides multi-granularity contextual information for distinguishing similar vertebrae.
%------------------------------------------------------------------------------------------
\section{Experiments}
\subsubsection{Datasets and Metrics.}
We evaluate the effectiveness of our approach on two public vertebrae segmentation datasets.
The VERSE2019 dataset~\cite{Sekuboyina} consists of 160 CT scans with 26 segmentation classes, covering vertebrae from the cervical to the lumbar regions.
The LUMBAR dataset~\cite{Khalil} contains 156 MRI scans from multiple scanner models and vendors, including T1-weighted, contrast-enhanced T1-weighted, and T2-weighted sequences.
It provides segmentation labels for lumbar vertebrae and intervertebral discs.
We use only the lumbar vertebrae labels and conduct experiments on T1-weighted images due to their superior quality.
Following the previous works~\cite{Sekuboyina,Xing}, we use two standard metrics to evaluate segmentation performance: Dice Similarity Coefficient (DSC) and 95th percentile of Hausdorff Distance (HD95).
%------------------------------------------------------
\begin{figure}[t]
\includegraphics[width=\textwidth]{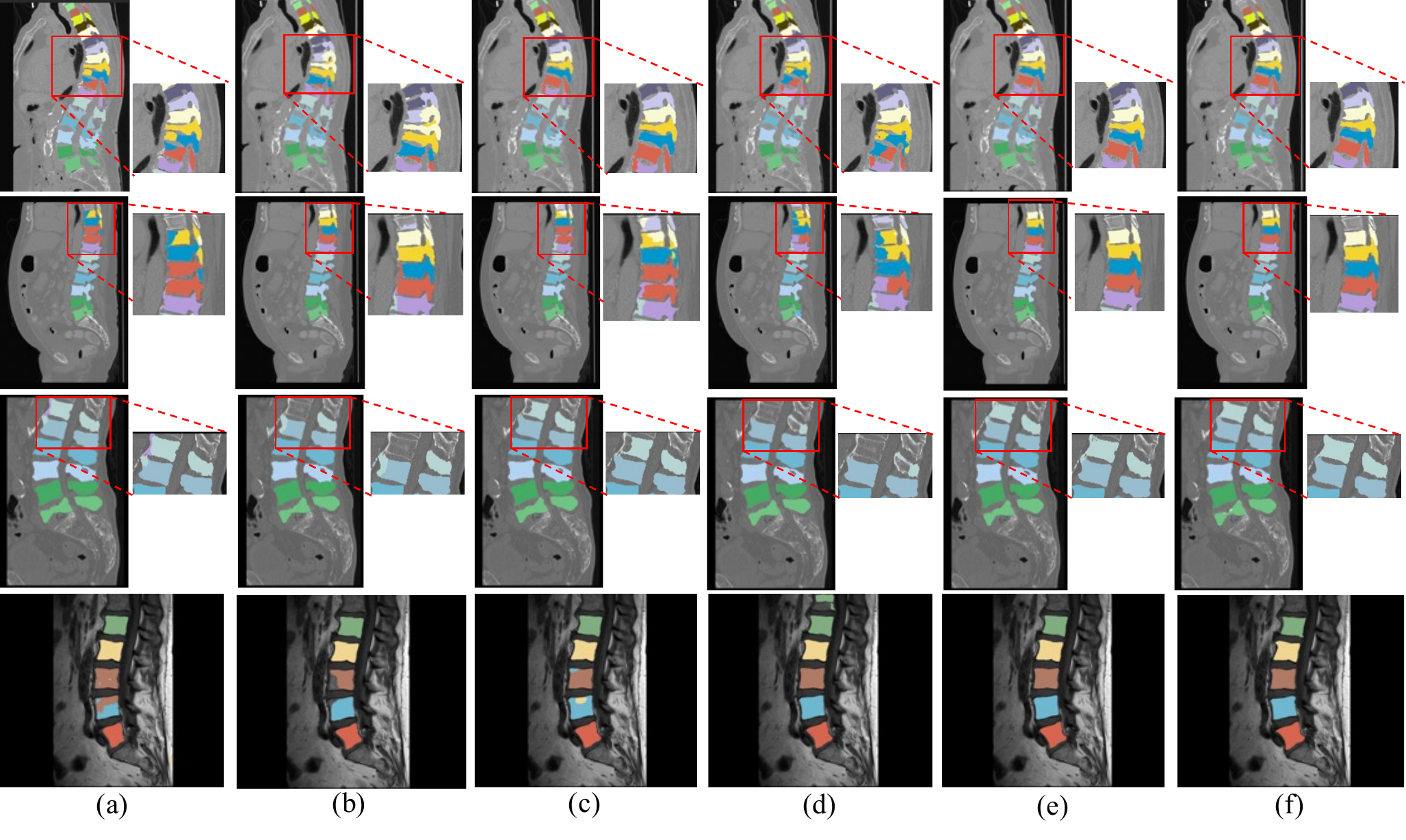}
\caption{Qualitative results with different methods. The top three rows display segmentation results from the VERSE2019 dataset, covering cervical, thoracic, and lumbar regions with various regions and views, whereas the bottom row shows results from the LUMBAR dataset. (a) nn-UNet, (b) UMamba, (c) MambaClnix, (d) Segmamba, (e) FMC-Net (Ours), (f) Ground Truth. }
\label{fig3}
\end{figure}
%-------------------------------------------------------

\subsubsection{Implementation Details.}
The proposed method is implemented in PyTorch and runs on an NVIDIA-A800-SXM4-80GB GPU.
It is integrated within the nnU-Net framework for automatic data processing and augmentation, and employs a combined loss of weighted cross-entropy and Dice losses at each decoder stage.
The model is trained on the VERSE2019 dataset with a patch size of (128, 160, 96) for 1000 epochs and on the LUMBAR dataset with a patch size of (32, 256, 160) for 600 epochs, both using a learning rate of 0.01.
%-----------------------------------------------------------
\begin{table}[t]
    \centering
    \caption{Comparison of different methods on VERSE2019 and LUMBAR datasets. Cer., Tho., Lum. denote cervical, thoracic, and lumbar vertebrae, respectively.}
    \scalebox{0.97}{
    % \begin{tabular}{lrrrrrrrrrr}
    \begin{tabular}{l *{10}{c}}
    \toprule
    \multirow{2}{*}{Methods} &
    \multicolumn{5}{c}{DSC (\%)} &
    \multicolumn{5}{c}{HD95 (mm) $\downarrow$} \\
    \cmidrule(lr){2-6} \cmidrule(lr){7-11}
    & \multicolumn{4}{c}{VERSE2019} & LUMBAR &
    \multicolumn{4}{c}{VERSE2019} & LUMBAR \\
    \cmidrule(lr){2-5} \cmidrule(lr){6-6} \cmidrule(lr){7-10} \cmidrule(lr){11-11}
    & Cer. & Tho. & Lum. & Mean & Lum. & Cer. & Tho. & Lum. & Mean & Lum. \\
    \midrule
    % 第一段：3DUNet 和 nn-UNet
    %3DUNet & - & - & - & - & 48.07 & - & - & - & - & 73.41 \\
    nn-UNet & 77.89 & 74.77 & 70.95 & 74.73 & 61.76 & $\textbf{1.21}$ & 5.12 & $\textbf{3.94}$ & 3.74 & 39.38 \\
    \cmidrule(lr){1-11}
    % 第二段：Cotr 和 Swin-UNetR
    CoTR & 74.56 & 67.15 & 64.57 & 68.61 & 72.55 & 7.93 & 10.15 & 11.39 & 9.83 & 31.87 \\
    Swin-UNetR & 79.30 & 56.90 & 60.90 & 64.73 & 47.75 & 2.12 & 17.15 & 9.07 & 11.00 & 97.28 \\
    \cmidrule(lr){1-11}
    % 第三段：UMamba、SegMamba、MambaClnix 和 Ours
    UMamba & 83.54 & 79.34 & 71.01 & 78.28 & 72.47 & 1.60 & $\textbf{3.13}$ & 4.75 & 3.09 & 27.59 \\
    SegMamba & 73.90 & 74.63 & 65.43 & 72.22 & 78.17 & 5.70 & 4.86 & 11.70 & 6.73 & 17.53 \\
    MambaClinix & 83.94 & 71.26 & 67.93 & 74.01 & 55.79 & 1.91 & 7.80 & 5.54 & 5.61 & 29.09 \\
    Ours & $\textbf{85.46}$ & $\textbf{80.74}$ & $\textbf{71.83}$ & $\textbf{79.92}$ & $\textbf{79.13}$ & 1.36 & 3.32 & 4.47 & $\textbf{3.04}$ & $\textbf{17.13}$ \\
    \bottomrule
    \end{tabular}}
    \label{tab:compare}
\end{table}
%-----------------------------------------------------------------------------
\begin{table}[t]
    \centering
    \caption{Ablation experiments with different components on LUMBAR dataset.}
    \begin{tabular}{c c c c c c}
        \toprule
        Baseline & DWT-Sample & HFR & MG-SSM & DSC\% & HD95 \\
        \midrule
        \checkmark &  &  &  & 71.09 & 42.77 \\
        \checkmark & \checkmark &  &  &  71.75 & 22.12 \\
        \checkmark & \checkmark & \checkmark &  & 75.30 & 24.52\\
        \checkmark & \checkmark &  & \checkmark & 76.56 & 18.67\\
        \checkmark &  &  & \checkmark & 75.94 & 16.72\\
        \checkmark & \checkmark & \checkmark & \checkmark & 79.13 & 17.13\\
        \bottomrule
    \end{tabular}
    \label{tab:ablation}
\end{table}
%-----------------------------------------------------------------------------

\subsubsection{Comparison with State-of-the-art Methods.}
We compare the proposed model with three categories of methods to evaluate its effectiveness.
The first category includes CNN-based methods, such as nn-UNet~\cite{Isensee}.
The second category includes Transformer-based approaches, comprising Swin-UNetR~\cite{Tang} and CoTR~\cite{Xie}.
The third category consists of Mamba-based methods, including UMamba~\cite{Ma}, SegMamba~\cite{Xing} and MambaClinix~\cite{Bian}.
he evaluation results on the VERSE2019 and LUMBAR datasets are presented in Tab.~\ref{tab:compare}.
The VERSE2019 dataset is very challenging due to the inclusion of cervical, thoracic, and lumbar regions with varying fields of view.
Although the performance improvements of our method over existing approaches may be modest in certain regions, it still achieves a new state-of-the-art.
Compared with the best-performing Mamba-based method, SegMamba, our approach improves the mean DSC per class by 1.64\%.
On the LUMBAR dataset, which contains only lumbar vertebrae, our method demonstrates a more substantial performance gain, achieving 79.13\% in DSC and 17.13 in HD95.
Fig.~\ref{fig3} shows representative visualizations from both datasets for qualitative comparison.
The visualizations demonstrate that our method has a significant advantage in distinguishing similar vertebrae, showcasing its robustness and effectiveness in the vertebrae segmentation task.

\subsubsection{Ablation Study.}
To evaluate the effectiveness of each component within the proposed model, we conduct ablation studies on the LUMBAR dataset.
Tab.~\ref{tab:ablation} illustrates the contribution of each component to the performance.
Specifically, we employ a standard U-shaped model with additive residual blocks as the baseline.
Subsequently, we incrementally integrate the DWT-Sample, HFR, and MG-SSM to evaluate the effectiveness of each component.
Here, DWT-Sample refers to the combination of WTD and WTU, which are utilized for downsampling and upsampling during the encoder and decoder stages, respectively.
The results demonstrate that each component significantly enhances the overall performance, confirming the effectiveness of each individual component.
%----------------------------------------------------------------------
\section{Conclusion}
In this paper, we propose a novel vertebrae segmentation network, FMC-Net, which effectively addresses the challenges of image blurring and the difficulty in distinguishing similar vertebrae.
We take advantage of the wavelet transform to prevent the information loss, decomposing the features into low-frequency and high-frequency components for feature enhancement.
HFR refines the high-frequency components, successfully restoring fine-grained details and alleviating image blurring.
MG-SSM extracts spatially-varying contexts and long-range dependencies, which enable the full utilization of the rich information contained in the low-frequency components.
Extensive experiments demonstrate that FMC-Net outperforms state-of-the-art approaches on both CT and MRI vertebrae segmentation datasets.
%-------------------------------------------------------------------------------
\begin{credits}
\subsubsection{\ackname} This study was supported by the Fundamental Research Funds for the Central Universities with No. DUT24YG135 and DUT23YG232.
\end{credits}

\subsubsection{\discintname}
All authors declare that they have no conflicts of interest.
% \subsubsection{\discintname}
% It is now necessary to declare any competing interests or to specifically
% state that the authors have no competing interests. Please place the
% statement with a bold run-in heading in small font size beneath the
% (optional) acknowledgments\footnote{If EquinOCS, our proceedings submission
% system, is used, then the disclaimer can be provided directly in the system.},
% for example: The authors have no competing interests to declare that are
% relevant to the content of this article. Or: Author A has received research
% grants from Company W. Author B has received a speaker honorarium from
% Company X and owns stock in Company Y. Author C is a member of committee Z.
% \end{credits}

% \end{comment}

\end{document}